\documentclass[final,5p,times,twocolumn]{elsarticle}

\usepackage{graphicx}
\usepackage{amssymb}
\usepackage{amsthm}

\usepackage{lineno}

\usepackage{amsmath}
\usepackage{multirow}
\usepackage{subfig}
\usepackage{float}
\usepackage{amsmath}
\newcommand{\abs}[1]{\left\lvert #1 \right\rvert}

\usepackage{romannum}
\usepackage{color,soul}
\usepackage{algorithm}
\usepackage{algorithmic}
\newcommand{\transpose}[1]{\ensuremath{#1^{\scriptscriptstyle T}}} 

\journal{xxxx}

\begin{document}

\begin{frontmatter}


\title{TASAC: a twin-actor reinforcement learning framework with stochastic policy for batch process control }

%% \fntext[label2yeer5eq4q]{}

\author{Tanuja Joshi$^a$}
\author{Hariprasad Kodamana$^{a,b,}$\fnref{label2}}
\ead{kodamana@iitd.ac.in}
\fntext[label2]{Corresponding authors}
\author{Harikumar Kandath$^{c,}$\fnref{label2}}
\ead{harikumar.k@iiit.ac.in}
\author{Niket Kaisare$^d$}
\address{$^a$Department of Chemical Engineering,
Indian Institute of Technology Delhi,
Hauz Khas, New Delhi - 110016\\
$^b$School of Artificial Intelligence,
Indian Institute of Technology Delhi,
Hauz Khas, New Delhi - 110016\\
$^c$International Institute of Information Technology Hyderabad,
Gachibowli, Hyderabad - 500 032, India\\
$^d$ Department of Chemical Engineering,
Indian Institute of Technology Madras, Chennai-600036, India}

\begin{abstract}
Due to their complex nonlinear dynamics and batch-to-batch variability, batch processes pose a challenge for process control. Due to the absence of accurate models and resulting plant-model mismatch, these problems become harder to address for advanced model-based control strategies. Reinforcement Learning (RL), wherein an agent learns the policy by directly interacting with the environment, offers a potential alternative in this context. RL frameworks with actor-critic architecture have recently become popular for controlling systems where state and action spaces are continuous. It has been shown that an ensemble of actor and critic networks further helps the agent learn better policies due to the enhanced exploration due to simultaneous policy learning. To this end, the current study proposes a stochastic actor-critic RL algorithm, termed Twin Actor Soft Actor-Critic (TASAC),  by incorporating an ensemble of actors for learning, in a maximum entropy framework, for batch process control. \end{abstract}


\end{frontmatter}

%\linenumbers
\allowdisplaybreaks
\section{Introduction}
\label{S:1}
Batch processes are widely used for the production of value-added products and specialty chemicals. Contrary to the continuous processes, batch processes have the advantage of flexibility in operations, low capital and raw material cost, and options to have a wider product range. The product quality obtained at the end of the batch process is of utmost importance. However, due to the inherent non-linearity, time-varying dynamics, batch-to-batch variations which are unique to batch processes, etc.,  optimization and control of batch processes are challenging tasks. 

Model-based control approaches are popularly being used for batch process control \cite{chen2012deterministic,lee2000convergence,hariprasad2016computationally,hariprasad2016efficient,mate2019stabilizing}. Model predictive control is an advanced control strategy extensively used in the process industries \cite{Kern2015AdvancedReactor,De2016DynamicProduction} and employs  mathematical programming to solve a constrained, possibly non-convex, optimisation problem. The key issue here is that online computation required at each time step to obtain the optimal control input profile is very high. This limits the implementation of these approaches for complex nonlinear, high-dimensional dynamical systems, despite the advancement in computational hardware and numerical methods. Further, the performance of a model-based controller is highly dependent on the availability of the accurate process model. For a complex and nonlinear process, the availability of such a model is a limitation as it requires significant prior knowledge and expertise. Plant-model mismatch occurs even if there is a slight variation in the real process and its approximate model, leading to inaccurate predictions of the performance variable, such as product yield. Even if the process model is available, the controller performance deteriorates in the presence of uncertainties and process drifts.  Batch-to-batch variations that occur due to raw material fluctuations, cleaning, etc., are a source of uncertainty that further deteriorate the closed loop performance. To overcome some of the difficulties
in generating accurate first-principle based models, data-driven modelling approaches for  batch processes having non-linear dynamics are reported in
the literature \cite{joshi2020novel,jiang2019data}.

In this juncture, the development of a control strategy that does not rely on  the  knowledge of the accurate dynamics of the system and can handle the stochastic dynamics and plant-model mismatches is extremely useful. To address these challenges, Reinforcement Learning (RL) based control, as a data-driven approach, is proven to be a good candidate as an alternative to traditional control approaches \cite{yoo2021reinforcement,bao2021deep,oh2022integration}. Contrary to traditional controllers, in RL based control, the agent (analogous to the controller) learns the optimal control action  (analogous to control input) by directly interacting with  the environment (analogous to the system/process)  through a data driven approach \cite{sutton2018reinforcement}. %The data can be real plant data or simulated data and contains(or made to contain) the effect of uncertainties. 
RL agent continuously improves the policy by means of the reward that is earned through these repeated interactions. This kind of model-free RL approach addresses the main limitation of model-based control approaches by eliminating the requirement of a high fidelity process model. Even if an approximate process model is available, it can be used in the offline learning stage and data generation \cite{joshi2021twin}, thereby significantly reducing the  requirement of data and the risk associated with safety. Thus, the burden on the online computation will be relatively less as the policy obtained through offline learning can be used as the warm start during the online implementation. 

Possibly owing to its simplicity, Q-learning has been a commonly used algorithm in the RL domain \cite{lee2005approximate}. In Q-learning, the optimal 'cost-to-go' function is learned in an iterative manner via value-iteration to find the optimal policy. Value iteration has its roots in dynamic programming, where it is the widely used algorithm. `Q-learning' is an Approximate Dynamic Programming (ADP) approach where the optimal policy is learned by value iteration in a model-free manner \cite{lee2005approximate,lee2006choice,peroni2005optimal}. In the traditional Q-learning framework, both state and action space are considered to be discrete. However, for process control applications,  where both the state space and the action space are continuous and high dimensional, the application of traditional Q-learning is limited. To overcome this, deep Q-learning has been introduced by Mnih et al., wherein the neural networks are used as function approximators for approximating the Q-values \cite{mnih2015human}. Some of the current works on RL applications in chemical processes employing Q-learning and deep Q-learning are related to applications such as polymerization and  chromatography  \cite{nikita2021reinforcement,singh2020reinforcement}. 
Another approach for solving an RL problem is known as policy gradients which directly optimizes a policy without using a value function or Q-value explicitly.
It uses a parameterized policy and a gradient ascent optimizer to maximize the expected rewards. Unlike Q-learning, policy-based methods can learn stochastic policies and are effective in continuous and high dimensional action spaces. For example, Petsagkourakis et al. have applied the policy gradient approach to find the optimal policy for complex and uncertain batch bio-processes \cite{Petsagkourakis2020ReinforcementOptimization}. 

 Actor-Critic algorithms combine both the value-based and the policy-based approach by learning estimates of both policy and value functions. It is the widely used framework applicable for dealing with  problems involving continuous action spaces. Deep deterministic policy gradient (DDPG) \cite{lillicrap2015continuous} is a
widely-used RL algorithm for continuous control that is built upon the actor-critic framework. DDPG is well-known for its effectiveness in problems involving continuous states and action spaces, such as chemical process control.
 For example, Yoo et al. has proposed a Monte Carlo-DDPG based controller for batch polymerisation control with emphasis on the reward design to satisfy both the economic and path/end point constraint and multiple phases in batch processes and showed enhanced ability than NMPC \cite{yoo2021reinforcement}. Ma et al. developed a DDPG based controller for semi-batch polymerisation system \cite{Ma2019ContinuousLearning}. The references provide a detailed review on the application of RL in the process control domain. \cite{shin2019reinforcement,spielberg2019toward,nian2020review}. Although DDPG has been widely used for continuous control, however, there are some limitations which include the overestimation of bias,  sensitivity to hyper-parameters, etc., which limits its implementation. 

It has been shown that  RL algorithms such as proximal policy optimization (PPO), TRPO and    DDPG, TD3, etc.,\cite{fujimoto2018addressing,lillicrap2015continuous,schulman2017proximal}  suffers from high sample complexity and brittle convergence, respectively. To address this challenge Soft actor critic (SAC), an off-policy actor-critic algorithm, has been introduced by Haarnoja et al. in a  maximum entropy framework \cite{haarnoja2018soft}. SAC incorporates an entropy-based RL objective and two critic networks for addressing overestimation bias. SAC uses a stochastic policy which enhances the stability with more stochastic exploration and faster convergence, yielding robust policies. 
Recently, SAC algorithms have been shown to have applications in building energy management \cite{Coraci2021}, energy management of  Electric vehicles with a hybrid energy storage systems\cite{xu2022soft}. Zhang et al. have proposed a SAC based control algorithm for minimizing the operational costs and ensuring power reliability in an integrated power, heat and natural-gas system \cite{zhang2021soft}.
However, the use of the SAC,  as a potential RL algorithm,  in the process control domain,  has not been reported in the literature \cite{liu2020reinforcement,Chen2020,Lyu2021,joshi2021twin}

The aforementioned algorithms updates the policy based on the gradient ascent approach and, therefore, can easily trap in the local optima rather than global optima. Recently, there has been an interest in the research fraternity to employ an ensemble of actor networks for improving the performance of the RL agents \cite{Lyu2021,huang2017learning,zhang2019ace,liu2020reinforcement,joshi2021twin}. The idea is that having multiple actors will allow the agent to explore different paths instead of getting restricted to a single path if it would have only one policy.
To this end, this study proposes a stochastic actor-critic RL algorithm termed Twin actor Soft Actor Critic (TASAC) combining SAC and multi-actor frameworks.
The efficacy of the TASAC algorithm is validated by comparing it with the  SAC algorithm for the control of batch transesterification process. 
We showcase that TASAC algorithm shows better performance in the presence of measurement noise and batch-to-batch variations as compared to vanilla SAC  and DDPG, which has a deterministic policy. The major contributions of this research are (i) SAC based control has not been studied much in the process control domain, hence this research is conducive to enrich the literature of SAC based batch process control ii) amalgamation of twin-actor framework in SAC for performance improvement, yielding an RL algorithm termed as twin actor SAC (TASAC)   has not been reported in the literature of process control, iii)  comparison with vanilla SAC  and DDPG algorithms for the control  of batch transesterification process,  showing that TASAC performs well, particularly in handling uncertain dynamics and measurement noise.

The remainder of the paper is divided into the following sections. The preliminaries/background required for developing a TASAC controller is presented in Section 2. Section 3 explains the TASAC algorithm in detail. Section 4 discusses the results where TASAC algorithm is applied for the control of batch transesterification process. The study's concluding remarks are summarised in Section 5.

\section{Background}
\subsection{Reinforcement Learning}
RL is a class of machine learning paradigm wherein feedback from the environment is rewarded based on the desired behaviour.
The framework of RL consists of an agent, environment, and a reward function \cite{sutton2018reinforcement}. The policy, $\pi$ is a function that maps the state $s$ to a single action $a$ if it is a deterministic policy, $  a \sim \pi(s)$  or a distribution of action if it is a stochastic policy, $ a \sim p(a|s) $ \cite{sutton1999policy} .
The RL agent outputs the action $a_{t}$ based on the policy $\pi$, upon receiving state $s_{t}$ from the environment resulting in a reward $r_t \coloneqq r(s_t,a_t)$. The new action transitions the environment to the next state $s_{t+1}$  resulting  a sequence of   state, action, reward and next state tuples $(s_{t},a_{t},r_{t},s_{t+1})$.
The goal in RL is to obtain a policy, $\pi^{*}$, that maximises the expected discounted reward as given below:

\begin{equation}
%\label{reward1}
\pi^{*} =  \arg \underset{\mathcal{\pi}}{ \max}\,\,\mathbb{E}~[R_{t} :=\sum\limits_{k=0}^\infty   \gamma^{k}{r_{t+k}}]\label{eq:rl}
\end{equation}
where $r_{t+k}$ is the reward at k steps ahead, and $\gamma$ is a suitable discount factor. There are two basic frameworks to train an RL agent:  value-based approach and policy-based approach. In the value-based approach, the Q-value of a state-action pair for policy $\pi$, $Q^{\pi}(s,a)$, is defined as the expected return when the agent starts at a state $s$ and performs an action $a$ following
a policy $\pi: \mathcal{S} \rightarrow \mathcal{A}$ as given below:
\begin{equation}
    Q^{\pi}(s,a)= \mathbb{E}_{\pi}\Big[\sum\limits_{k=0}^\infty   \gamma^{k}{r_{t+k}}|s_{t}=s,a_{t}=a\Big] 
    \label{Qvalue}
\end{equation}
Value-based approach aims to learn the optimal Q-value, $Q^{*}(s,a)$, which is the maximum expected return achieved by the optimal policy for the state-action pair, in order to find the optimal policy, $\pi*$.
%Value-based approach aims to find the optimal policy, $\pi^{*}$, by learning the optimal Q-value, $Q^{*}(s,a)$, which is the maximum expected return achieved by the optimal policy for the state-action pair. 
The optimal action ($a^*$) in the value based approach is evaluated as:
\begin{equation}
a^* =  \arg \underset{a} {\max} Q^{\pi}(s,a)
\end{equation}

%(as given in eqn \eqref{Qvalue}) 
%\subsection{Policy Gradient Methods}
In the policy gradient methods,  the objective is to maximise the expected discounted reward defined as $ J(\pi_{\phi_{\mathbf{A}}})=\mathop{\mathbb{E}}_{\tau \sim \pi_{\phi_{\mathbf{A}}}}[R(\tau)] $  to learn an optimal policy parameterized by $\theta$. Here, $R(\tau)$ denotes the return obtained from the trajectory $\tau = \{s_0,a_0,s_1,a_1,\dots\}$
  Unlike the value-based methods, which learns an approximate function for the optimal Q-value, $Q^{*}(s,a)$, to find the optimal policy, policy gradient methods directly optimise the policy (defined explicitly) without using a value function and are suitable for continuous and stochastic environments.
Policy gradient methods aim to find the parameter $\theta$ that maximises this objective function using gradient ascent \cite{sutton1999policy}. 
Mathematically, the policy-gradient approach is given as:
\begin{equation}
    \displaystyle \nabla_{\phi_{\mathbf{A}}} (J(\pi_{\phi_{\mathbf{A}}}))=\mathop{\mathbb{E}}_{a \sim \pi_{\phi_{\mathbf{A}}}}\left[\nabla_{\phi_{\mathbf{A}}}\log \pi_{\phi_{\mathbf{A}}}(a_{t}|s_{t})Q^{\pi}(s,a)\right]
\end{equation}
where $\pi_{\phi_{\mathbf{A}}}(a_{t}|s_{t}) = Pr \{a_t | s_t\}$ is the probability that action a is taken at time t given that the environment is in state s at time t with parameter $\phi_{\mathbf{A}} $.
This results in the following update equation for $\phi_{\mathbf{A}}$:
\begin{equation}
    \phi_{\mathbf{A}_{m+1}} = \phi_{\mathbf{A}_{m}} + \beta_{\phi_A}\displaystyle \nabla_{\phi_{\mathbf{A}}} (J(\pi_{\phi_{\mathbf{A}}}))
\end{equation}
where $m$ is the iteration index and $\beta$ is the learning rate. 
%\subsection{Actor-critic RL algorithms}

Policy-based (actor) techniques tend to be more stable, but they have the disadvantage of having high variance estimations of the gradient, which results in slow convergence. Value-based (critic) approaches are a more indirect method of policy optimization, but they are more sample efficient. Actor-critic methods are a class of algorithms that provides `the best of both worlds by learning estimates of both value and policy functions \cite{degris2012off,NIPS1999_6449f44a}.

Actor-critic algorithms estimates the Q-value, $ Q^{\pi}(s,a)$,  using a function approximator termed `critic' to yield approximate Q-value, $ Q_{\phi_{\mathbf{C}}}(s,a)$ parameterised with the parameter $\phi_{\mathbf{C}}$. The policy (actor) updates the parameter $\phi_{\mathbf{A}}$ in a direction as suggested by the critic. Further,  the critic evaluates the action output by the actor and updates the Q-value function parameter $\phi_{\mathbf{C}}$. The  actor learns by penalising   the approximate policy gradient as given below:
\begin{equation}
    \displaystyle \nabla_{\phi_{\mathbf{A}}} (J(\pi_{\phi_{\mathbf{A}}})) \approx \mathop{\mathbb{E}}_{a \sim \pi_{\phi_{\mathbf{A}}}}\left[\nabla_{\phi_{\mathbf{A}}}\log \pi_{\phi_{\mathbf{A}}}(a_{t}|s_{t})Q_{\phi_{\mathbf{C}}}(s,a)\right]
\end{equation}
The actor parameters, $\phi_{\mathbf{A}}$, are updated using the following update equation:
\begin{equation}
\label{actor_update}
    \Delta \phi_{\mathbf{A}} = \beta_{\phi_{\mathbf{A}}} \nabla_{\phi_{\mathbf{A}}}\log \pi_{\phi_{\mathbf{A}}}(a_{t}|s_{t}) *Q_{\phi_{\mathbf{C}}}(s_{t},a_{t})  
\end{equation}

The critic parameter, $\phi_{\mathbf{C}}$ is updated by minimising the temporal differnece (TD) error using the following update equation:
\begin{equation}
\label{critic_update}
    \Delta \phi_{\mathbf{C}} = \beta_{\phi_{\mathbf{C}}}\left (r(s_{t},a_{t}) +\gamma Q_{\phi_{\mathbf{C}}}(s_{t+1},a_{t+1})-Q_{\phi_{\mathbf{C}}}(s_{t},a_{t})\right)
    \nabla_{\phi_{\mathbf{C}}}(Q(s_{t},a_{t})) 
\end{equation}

where $ \delta = \big[r(s_{t},a_{t}) +\gamma Q_{\phi_{\mathbf{C}}}(s_{t+1},a_{t+1})\big]-Q_{\phi_{\mathbf{C}}}(s_{t},a_{t})$  is the TD error. Here the term in the square bracket is the target value at time step $t+1$ formed by the immediate reward, $r_t$, and next time step estimate, $Q_{\phi_{\mathbf{C}}}(s_{t+1},a_{t+1})$. Thus, the goal is to minimise the difference between the target value at time $t+1$ and the estimate at current time step $t$.
The parameter updates in equations \ref{actor_update} and \ref{critic_update} together constitutes actor-critic updates of RL.

RL algorithms can be broadly classified on how the optimal policy is learned, namely,  on-policy and off-policy learning. Algorithms such as TRPO, PPO, etc. \cite{schulman2017proximal} are based on-policy learning and therefore suffers from high sample complexity. To address this problem, off policy-based algorithms such as DDPG, TD3 \cite{lillicrap2015continuous,fujimoto2018addressing} were introduced that used a replay buffer to store the past transitions, and the optimal policy is obtained by exploiting the past experiences. However, these algorithms require meticulous hyperparameter tuning and thus leading to brittle convergence. 

% \subsection{DDPG}
% DDPG has been widely used actor-critic RL algorithm for continuous control. DDPG algorithm learns a deterministic policy by incorporating the merits of Deep-Q Networks and replay buffer to optimize the Deterministic policy gradient(DPG) objective i.e $\nabla_{\theta} J(\mu_{\theta}) = \displaystyle \mathop{\mathbb{E}}[\nabla_{\theta}Q_{\mu}(s,a)|_{a=\mu_{\theta}(s)}]$. However the overstimation bias due to function approximation error and hyperparameter sensitivity makes its implementation limited in real application

\subsection{Maximum entropy RL and SAC}
Traditional RL algorithms aim to achieve the optimal policy by maximising the expected reward. This fundamentally discourages exploration as it will inhibit the agent to search for random policies by exploiting the maximum reward policy. Entropy in RL refers to the randomness of actions, and thus entropy-based RL tries to maximise the expected entropy of the policy in addition to the expected return. Thus, the learning will be more useful when the agent gets into unexplored regions in the environment during the testing phase. The policy achieved in maximum entropy RL is more adaptive as compared to traditional RL. Mathematically the objective of maximum entropy RL is given as:
\begin{equation}
\label{MERL_obj}
    J(\pi)= \mathbb{E}\Big[ \sum\limits_{t=0}^T \gamma^{t}\Big( r(s_{t},a_{t}) + \alpha {H}(\pi(.|s_{t})\Big) \Big]
    \end{equation}
    and
    \begin{equation}
    \pi^{*} =  \arg \underset{\mathcal{\pi}}{ \max}\,\,\, \mathbb{E}\Big[ \sum\limits_{t=0}^T \gamma^{t}\Big( r(s_{t},a_{t}) + \alpha {H}(\pi(.|s_{t})\Big) \Big]
    \end{equation}
    where $ {H}= \mathbb{E}[-log \pi(a|s)] $ and $\alpha $ is the temperature parameter of the policy.

%\subsection{SAC}
SAC is an off-policy actor-critic RL algorithm that is based on the maximum entropy RL framework \cite{haarnoja2018soft}. It maximizes the entropy RL objective (equation \ref{MERL_obj}) by parametrizing a stochastic policy. Harnooja et al. proposed the SAC algorithm and showed that SAC outperforms the prior state-of-art algorithms and is more stable.
The SAC framework contains an actor network, $\pi_{\phi_{\mathbf{A}}}$ and two critic networks,  $Q_{\phi_{\mathbf{C}_j}},~j\in \{1,2\}$  to simultaneously learn the system dynamics. The critic networks are updated by minimizing the mean squared bellman error given as:

\begin{equation}
   % Critic~ Loss =  \mathbb{E}\Big[Q_{\phi_{\mathbf{C}_j}}(s,a)- r +\gamma\Big( \underset{j}  \min Q_{\phi_{\mathbf{C}_j,\mathbf{T}}}(s',\tilde{a_1})- \alpha \log(\pi_{\phi_{A}}(s'))\Big) \Big]
    Critic ~ Loss =  \mathbb{E}_{(s,a,r,s')\sim E}\Big[\Big(Q_{\phi_{\mathbf{C}_j}}(s,a)- y(s',r)\Big)^{2}\Big] ~j\in \{1,2\}
    \end{equation}
where $y(s',r)= r +\gamma\Big( \underset{j}  \min Q_{\phi_{\mathbf{C}_j,\mathbf{T}}}(s',\tilde{a})- \alpha \log\pi_{\phi_\mathbf{A}}(\tilde{a}|s')\Big)$
where $r$ is the immediate reward and $y(s',r)$ is the target value and $Q_{\phi_{\mathbf{C}_j,\mathbf{T}}}$ is the target Q-value.
The policy aims to maximise the sum of the expected Q-value and the expected future entropy as given below:
\begin{align}
    Actor~Loss &= \mathbb{E}_{a\sim \pi_{\theta}}\Big[\underset{j}\min Q_{\phi_{\mathbf{C}_j}}(s,a) - \alpha \log\pi_{\phi_\mathbf{A}}(a|s)\Big] \\
    &= \mathbb{E}_{\xi \sim \mathcal{N} }\Big[\underset{j}  \min Q_{\phi_{\mathbf{C}_j}}(s,\tilde{a}_{\phi_{\mathbf{A}}}(s,\xi)) - \alpha \log\pi_{\phi_\mathbf{A}}(\tilde{a}_{\phi_{\mathbf{A}}}(s,\xi))|s)\Big]
\end{align}
where $ \tilde{a}_{\phi_\mathbf{A}} = \tanh \big(\mu_{\phi_\mathbf{A}}(s)+ \sigma_{\theta}(s). \xi\big) $ and $\xi \sim \mathcal{N}(0,1)$.

Recent works in the literature have explored the deployment of an ensemble of actor networks in the actor-critic RL framework \cite{joshi2021twin}.
In complex environments where the best strategy cannot be represented by a single network, multiple actor networks can be used as a potential solution to learn the optimal policy.
This work combines the 'actor and critic ensemble' in the SAC algorithm for improving the policies by the parallel training of the actor networks. Each actor will be trained separately, and the best action is chosen based on the Q-value estimate from the action given by the each actor. 

\section {The proposed framework : TASAC}
\label{S:3}
\begin{figure*}
     \centering
     \includegraphics[width=\linewidth]{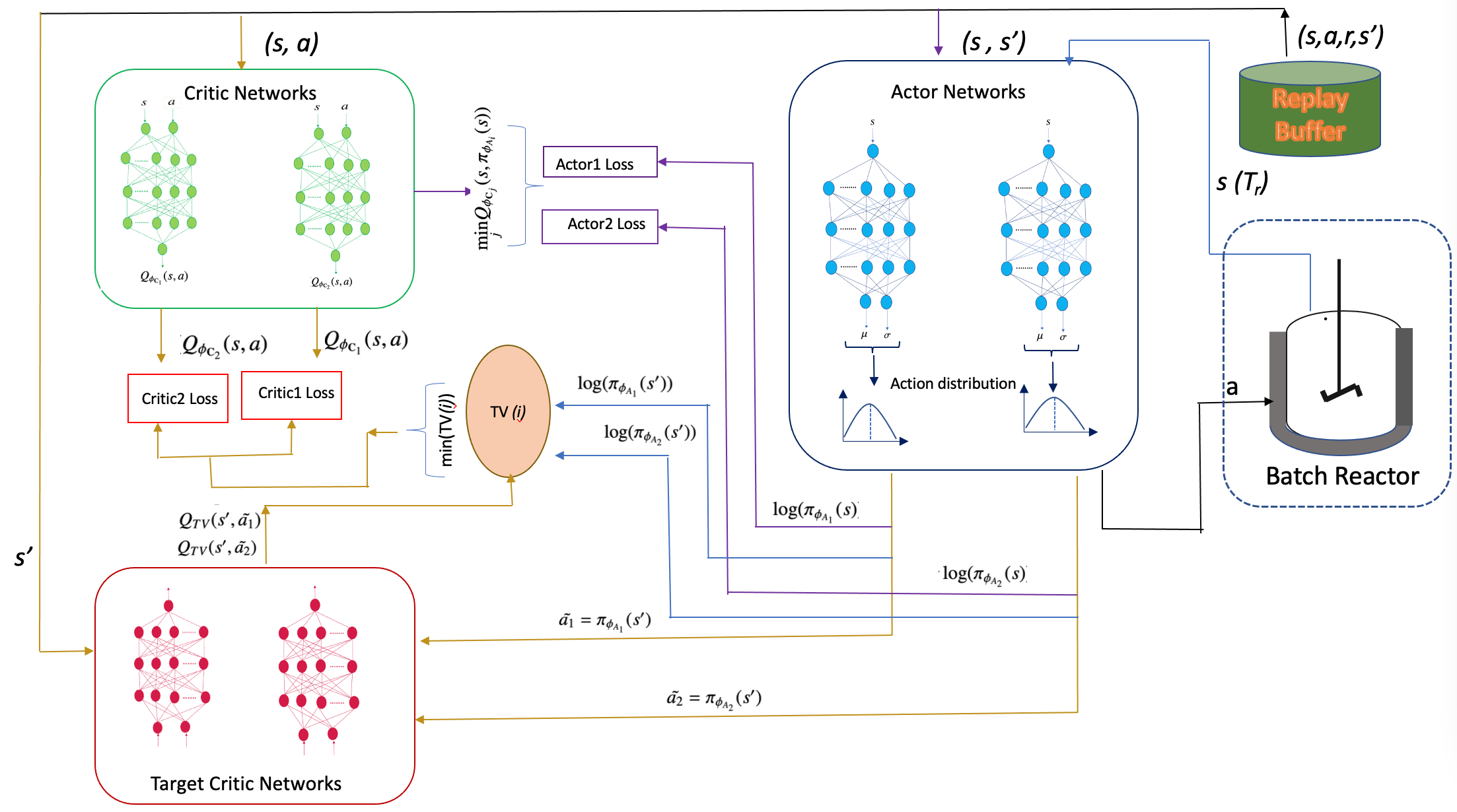}
     
    \caption{Schematic  of the TASAC based controller for batch process}
    \label{fig:Schematic}
\end{figure*}

This section presents the detail of the proposed TASAC algorithm.
TASAC consists of twin actors, in contrast to a single actor network as in the vanilla SAC algorithm, to achieve an optimal policy. The schematic of the TASAC based controller is shown in Figure {\ref{fig:Schematic}}. We propose to use two actor networks, $\pi_{\phi_{A1}}$ and $\pi_{\phi_{A2}}$, namely, with parameters $\phi_{A1}$ and $\phi_{A2}$, respectively. Overall TASAC algorithm consists of 6 Neural Networks (NN) , two actor networks , two critic networks  $Q_{\phi_{\mathbf{C}_{1}}}$, $Q_{\phi_{\mathbf{C}_{2}}}$ parametrised with $\phi_{\mathbf{C}_{1}}$,$\phi_{\mathbf{C}_{2}}$, respectively, and corresponding target critic networks, $Q_{\phi_{\mathbf{C}_{1},\mathbf{T}}}$ and $Q_{\phi_{\mathbf{C}_{2},\mathbf{T}}}$ with parameters, $\phi_{\mathbf{C}_{1},\mathbf{T}}$ and $\phi_{\mathbf{C}_{2},\mathbf{T}}$, respectively.

TASAC algorithm starts with the agent interacting with the environment and receives state $s$. Each actor  outputs a distribution of action corresponding to the same state $s$. Being stochastic policy generator,  action $a_{i}$ from the two actor networks is then sampled from the Normal distribution as given below:
\begin{equation}
a_i = \pi_{\phi_{\mathbf{A}_{i}}}(s) , ~~i \in 
       \{1,2\}
\end{equation}
The action  to be given to the environment is selected based on the Q-value estimate of the $(s,a_i)$ pair as given in equation
\begin{equation}
     a= \min\Big(  \underset{j}{\min} Q_{\phi_{\mathbf{C}_{j}}}(s,a_1),  \underset{j}{\min} Q_{\phi_{\mathbf{C}_{j}}}(s,a_2) \Big), ~~j \in \{1,2\}\label{a}
\end{equation}
Equation \eqref{a} indicates that for each action generated by the two actors, two Q-values are estimated and the minimum of them is selected to be given to the environment. This is based on the rational that if the Q-value is being overestimated, then taking the minimum of them will reduce the overestimation bias and leads to an optimal policy.

The learning starts by sampling a batch of transition tuples $(s, a, r, s')$ from the replay buffer.
Both the actor networks are fed with the next state $s'$ and they outputs the target action $\tilde{a_1}$ \& $\tilde{a_2}$ corresponding to the state, $s'$ i.e $\tilde{a_i}= \pi_{\phi_{A_i}}(s') ~~ i \in\{1,2\}$.  
The state $s'$ and target action $\tilde{a_i}$ is given as inputs to the target Q-network to estimate the target Q-value.
The target Q-values, $Q_{TV}(s,\tilde{a_1})$, $Q_{TV}(s,\tilde{a_2})$ for the state-action pairs $(s,\tilde{a_1})$, $(s, \tilde{a_2})$, respectively, are estimated by taking the minimum of the target values (TVs) of both the target Q networks:
\begin{equation}
    Q_{TV}(s',\tilde{a_1}) = \underset{j} \min \Big(Q_{\phi_{\mathbf{C}_j,\mathbf{T}}}(s',\tilde{a_1})\Big),~~j\in \{1,2\}
\end{equation}

\begin{equation}
    Q_{TV}(s',\tilde{a_2}) = \underset{j} \min \Big(Q_{\phi_{\mathbf{C}_j,\mathbf{T}}}(s',\tilde{a_2})\Big), ~~j\in \{1,2\}
\end{equation}
The TV to be used in the loss function of the critic is estimated by summing the current reward,  the discounted reward and the weighted entropy term.
\begin{equation}
    TV_{i}= \Big(Q_{TV}(s',\tilde{a_i})    - \alpha \log(\pi_{\phi_{A_{i}}}(s'))\Big), ~~i\in \{1,2\}
 \end{equation}                
 \begin{equation}
 \label{TV}
     TV= r+ \gamma  \underset{i}\min\Big(TV_i \Big) 
\end{equation}

TASAC algorithm computes the minimum of the two TVs obtained corresponding to the two target actions as given in equation \ref{TV} (Please refer to Step 12 of Algorithm1).
The critic networks calculate the Q-value of the $(s,a)$ pair sampled from the replay buffer. Finally, the critic networks are updated based on gradient descent of the mean square error (MSE) between the estimated Q-value, $Q_{\phi_{\mathbf{C}_{j}}} $ and the TV (Please refer to the Step 13 of Algorithm1).
\begin{equation}
    Critic 1 ~~Loss = MSE(Q_{\phi_{\mathbf{C}_1}}(s,a),TV) \\
\end{equation}
\begin{equation}
    Critic 2~~ Loss = MSE(Q_{\phi_{\mathbf{C}_2}}(s,a),TV) 
\end{equation}

 Finally, the actor networks are updated  to maximise the expected return and the entropy of the policy. This is done  by computing the  gradient ascent of the sum of the Q-value and  the entropy term as given below (Please refer to Step 14 of Algorithm 1):
\begin{equation}
    \underset{\phi_{A_{i}}}\max \mathbb{E}_{ \xi \sim \mathcal{N} }\Big[ \underset{j}\min Q_{\phi_{\mathbf{C}_{j}}}(s,\tilde{a}_{\phi_\mathbf{A}}(s,\xi))-\alpha \log(\pi_{\phi_{A_{i}}}(\tilde{a}_{\phi_\mathbf{A_i}}(s,\xi)  ))\Big] ,~~i\in \{1,2\}
\end{equation}
where $ \tilde{a
}_{\phi_\mathbf{A_i}} = \tanh \big(\mu_{\phi_{A_{i}}}(s)+ \sigma_{\phi_{A_{i}}}(s). \xi\big) $ and $\xi \sim \mathcal{N}(0,1)$.
Various steps described above have been concisely presented as Algorithm 1.

\subsection{Strategies for action selection and TVs}
\begin{figure*}
    \centering
    \includegraphics[width=\linewidth]{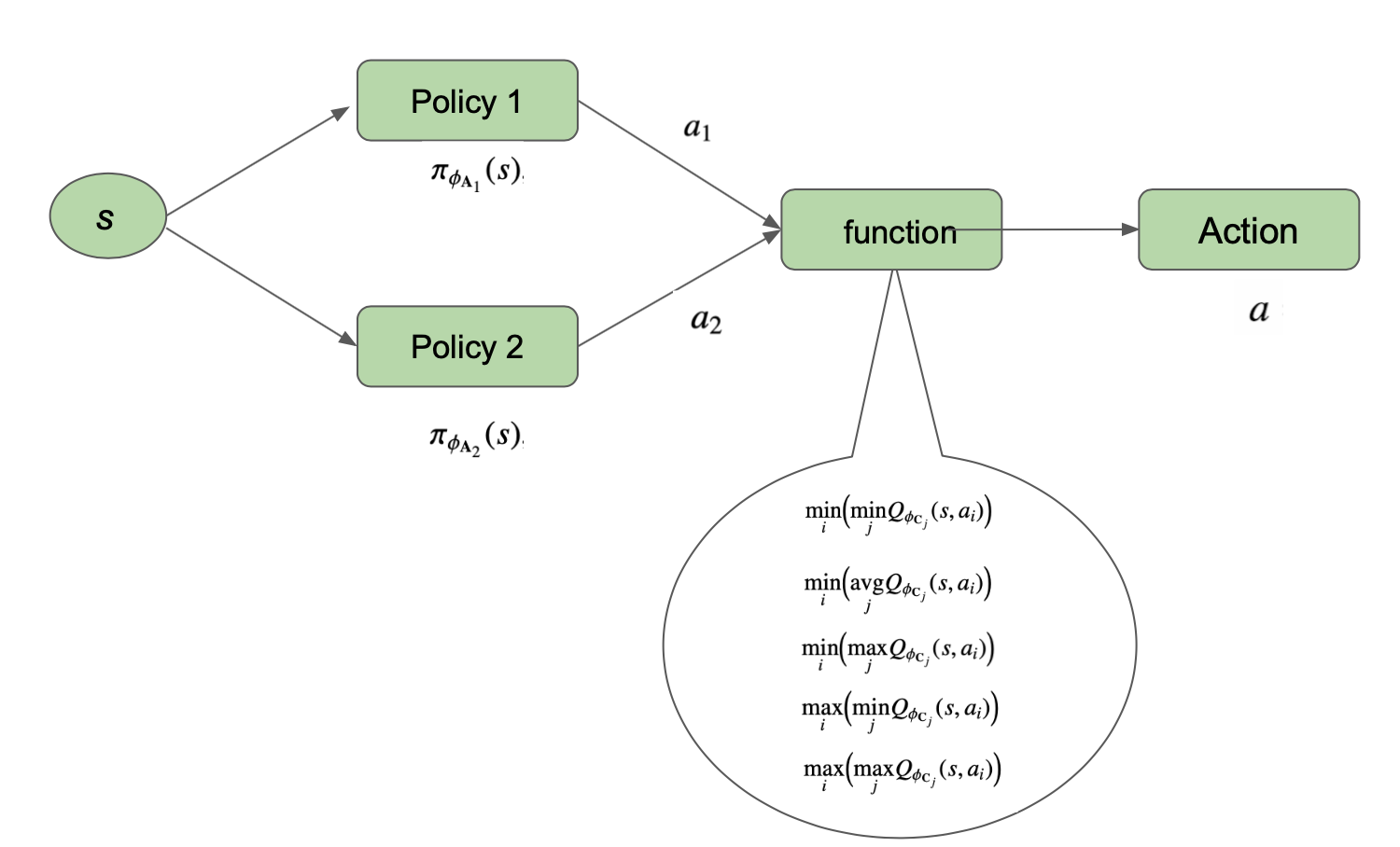}
    \caption{Selection of best action
    }
    \label{fig:select_action}
\end{figure*}

During the implementation of the  TASAC based controller, we have to choose the best action and TVs from the two actor networks.
%Multiple actor networks allow the agent to select the multiple actions, and the best action is then selected to be injected into the process. 
We propose five different functions to choose the best action and target values $(TV)$ from the ensemble of actor networks based on their Q-value estimates. The schematic for the same is shown in Figure \ref{fig:select_action}.

\begin{figure*}
    \centering
    \includegraphics[width=\linewidth]{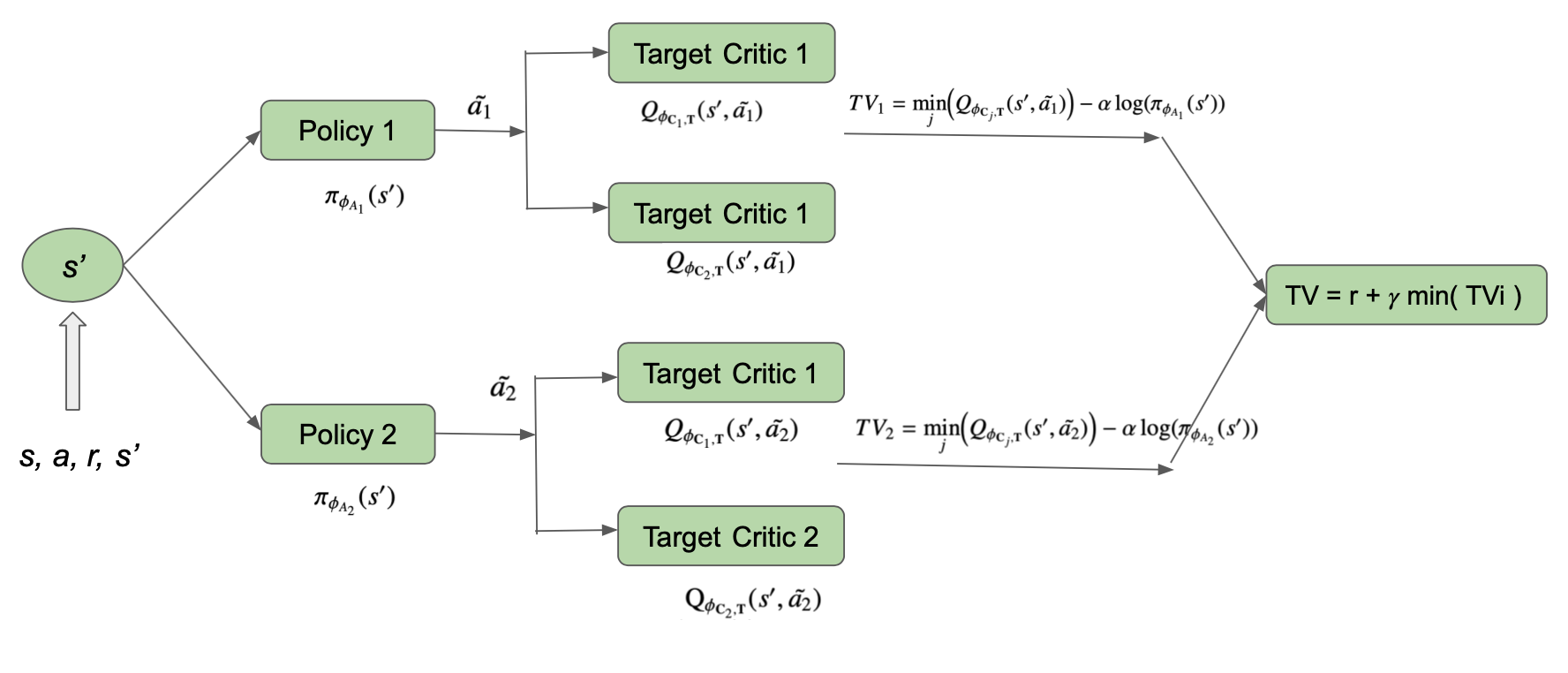}
    \caption{Selection of TV
    }
    \label{fig:select_TV}
\end{figure*}
\subsubsection{Selection of action}
\begin{itemize}
\item {\emph{Case 1: min-min}:} 
The actor network maps the current state to the action  as:
\begin{equation}
    a_i = \pi_{\phi_{\mathbf{A}_{i}}}(s) , ~~i \in 
       \{1,2\}
\end{equation}
where $a_i$ represents the action given by the $i^{th}$ actor. 
Initially,  each actor select the action based on the minimum of the two Q-value estimates of the state-action pair. Further, the action that minimizes the two output Q-values are chosen. This is mathematically represented  as below: 
\begin{equation}
\label{min_action}
     a= \underset{i}{\min}\Big(  \underset{j}{\text{min}} Q_{\phi_{\mathbf{C}_{j}}}(s,a_i) \Big),~i, j \in \{1,2\}
\end{equation}
The formulation is based on the rationale that if the Q-value is being overestimated by the NN, then evaluating the minimum will select the action that has a Q-value closer to the actual Q-value.

Suitably enumerating maximum (max), minimum (min), and average (avg), we have tried as different functions for the action selection. The simulation studies presented in the next section shows that the min-min case  performs the best among all the combinations. 

    \item Case 2: min-max : $ a = \underset{i}{\min}\Big(  \underset{j}{\max} Q_{\phi_{\mathbf{C}_{j}}}(s,a_i) \Big)$
%     \begin{equation*}
%     a = \underset{i}{\min}\Big(  \underset{j}{\max} Q_{\phi_{\mathbf{C}_{j}}}(s,a_i) \Big)
% \end{equation*}
    \item Case 3: max-min : $    a = \underset{i}{\max}\Big(  \underset{j}{\min} Q_{\phi_{\mathbf{C}_{j}}}(s,a_i) \Big)$
%     \begin{equation*}
%     a = \underset{i}{\max}\Big(  \underset{j}{\min} Q_{\phi_{\mathbf{C}_{j}}}(s,a_i) \Big)
% \end{equation*}
    \item Case 4: max-max :$ a = \underset{i}{\max}\Big(  \underset{j}{\max} Q_{\phi_{\mathbf{C}_{j}}}(s,a_i) \Big)$
%     \begin{equation*}
%     a = \underset{i}{\max}\Big(  \underset{j}{\max} Q_{\phi_{\mathbf{C}_{j}}}(s,a_i) \Big)
% \end{equation*}
\item Case 5: min-avg : $ a = \underset{i}{\min}\Big(  \underset{j}{\text{avg}} Q_{\phi_{\mathbf{C}_{j}}}(s,a_i) \Big)$
%     \begin{equation}
%     a = \underset{i}{\min}\Big(  \underset{j}{\text{avg}} Q_{\phi_{\mathbf{C}_{j}}}(s,a_i) \Big)
% \end{equation}
\end{itemize}
The schematic for this operation is shown in Figure \ref{fig:select_action}.
\subsubsection{Selection of Target Value (TV)} 
While calculating target Q-values for state $s'$  we need to employ two target action from each actor as shown in Figure \ref{fig:select_TV}:
\begin{equation}
    \tilde{a_i}= \pi_{\phi_{A_i}}(s') ~~ i \in\{1,2\}
\end{equation}
After enumerating all the five options as illustrated above, the function given in equation \eqref{min_action},  $ \underset{j}{\min} Q_{\phi_{\mathbf{C}_{j}}}(s,a_i) $,  is used to select the $TV_i$ i.e $\underset{j} \min \Big(Q_{\phi_{\mathbf{C}_j,\mathbf{T}}}(s',\tilde{a_i})\Big) - \alpha \log(\pi_{\phi_{A_{i}}}(s'))\Big)$, ~~$i,j\in \{1,2\} $  and then the minimum of the two $TVi$ is used in the calculation of TV given as:
$r+ \gamma  \underset{i}\min\Big(TV_i \Big)$ as it was yielding the best action.

%\item \emph{Case 2: min-avg}\\
%Here, first operation involves computing the  average of the two Q-values from each actor network  and the minimum of the output Q-values is evaluated, yielding, 
%\begin{equation}
    %\underset{i}{\min}\Big(  \underset{j}{\text{avg}} Q_{\phi_{\mathbf{C}_{j}}}(s,a_i) \Big)
%\end{equation}
%As discussed in Case 1, the calculation of the TV involves selecting the $TV_i$ by taking the average of the target Q-values for each target action from the two actors corresponding to state $s'$, that is, 
%$ \underset{j} {\text{avg}} \Big(Q_{\phi_{\mathbf{C}_j,\mathbf{T}}}(s',\tilde{a_i})\Big) - \alpha \log(\pi_{\phi_{A_{i}}}(s'))\Big)$, ~~$i,j\in \{1,2\} $ and then the minimum of the two $TV_i$ is used in the calculation of $TV$  as:
%$r+ \gamma  \underset{i}\min\Big(TV_i \Big)$. 

%\item \emph{Study3: min-max}
%\begin{equation*}
    %\underset{i}{\min}\Big(  \underset{j}{\max} Q_{\phi_{\mathbf{C}_{j}}}(s,a_i) \Big)
%\end{equation*}
%This function takes the max of the two Q values for the state-action pair, corresponding to each action value from the two actors i.e $\underset{j}{\max} Q_{\phi_{\mathbf{C}_{j}}}(s,a_i)$ and then the minimum of the Q values obtained from the first operator.

%\item \emph{Study4: max-min}
%\begin{equation*}
    %\underset{i}{\max}\Big(  \underset{j}{\min} Q_{\phi_{\mathbf{C}_{j}}}(s,a_i) \Big)
%\end{equation*}

%\item \emph{Study5: max-max}
%\begin{equation*}
    %\underset{i}{\max}\Big(  \underset{j}{\max} Q_{\phi_{\mathbf{C}_{j}}}(s,a_i) \Big)
%\end{equation*}
%\end{enumerate}

\begin{algorithm}
 \caption{TASAC Algorithm}
   \begin{algorithmic}[1]
   \STATE Initialise actor networks $\pi_{\phi_{\mathbf{A}_{i}}}$ $\&$ critic Networks $Q_{\phi_{\mathbf{C}_{j}}}$ with parameters $\phi_{\mathbf{A}_{i}} $, $\phi_{\mathbf{C}_{j}}, ~~~i,j \in 
       \{1,2\}$ 
    \STATE Initialise target critic networks $Q_{\phi_{\mathbf{C}_{j},\mathbf{T}}}$ with parameters, $\phi_{\mathbf{C}_{j},\mathbf{T}}~~~j \in 
       \{1,2\}$

     \FOR{$batch = 1 , no.~ of ~ batches$}
       \STATE Observe the initial state $s$
        \FOR{$time \in \{t_{init},\dots,t_{end}\}$}
          \STATE Compute action $a$, \\
           $a_i = \pi_{\phi_{\mathbf{A}_{i}}}(s) , ~~i \in 
       \{1,2\}$ \\
       $a= \min\Big(  \underset{j}{\min} Q_{\phi_{\mathbf{C}_{j}}}(s,a_1),  \underset{j}{\min} Q_{\phi_{\mathbf{C}_{j}}}(s,a_2) \Big),\,\, j \in \{1,2\}$
           
           \STATE Execute action $a$ to get $r$ and $s'$ 
            \STATE Add the tuple $(s,a,r,s')$ in replay memory E
             %\FOR {$j$ in range (iteration)}
               \STATE  Randomly sample a batch of transition $B$ from E 
                \STATE Compute the target action, ~$\tilde{a_i}= \pi_{\phi_{A_i}}(s') ~~ i \in\{1,2\}$
                 \STATE Compute the target Q-value :
                 
                 $Q_{TV}(s',\tilde{a_1}) = \underset{j} \min \Big(Q_{\phi_{\mathbf{C}_j,\mathbf{T}}}(s',\tilde{a_1})\Big)$ \\
                 $Q_{TV}(s',\tilde{a_2}) = \underset{j} \min \Big(Q_{\phi_{\mathbf{C}_j,\mathbf{T}}}(s',\tilde{a_2})\Big)$, ~~$j\in \{1,2\}$

                 \STATE Compute the targets: \\

                 $TV_{i}= \Big(Q_{TV}(s',\tilde{a_i})    - \alpha \log(\pi_{\phi_{A_{i}}}(s'))\Big)$, ~~$i\in \{1,2\}$

                 $TV= r+ \gamma  \underset{i}\min\Big(TV_i \Big)$

                  \STATE Update the critic networks:
                  
                 $\underset{\phi_{\mathbf{C}_{j}}} \min \Big(\frac{1}{\abs{B}} \sum(Q_{\phi_{\mathbf{C}_{j}}}(s,a)-TV)^{2}\Big) ~j\in \{1,2\}$
                  
                \STATE Update the actor networks:
                
               $\underset{\phi_{A_{i}}}\min \Big(\frac{1}{\abs{B}}\sum ( \alpha \log(\pi_{\phi_{A_{i}}}(s))- \underset{j}\min Q_{\phi_{\mathbf{C}_{j}}}(s,\pi_{\phi_{A_{i}}}(s)))\Big)$
               
               \STATE Update target critic networks.\\
               $\phi_{\mathbf{C}_j,\mathbf{T}} \leftarrow \tau  \phi_{\mathbf{C}_j} + (1-\tau)\phi_{\mathbf{C}_j,\mathbf{T}} ~~j\in \{1,2\}$ \\

               \STATE $s$ $\rightarrow$ $s'$  
              \ENDFOR
              
    \ENDFOR
  \end{algorithmic}
\end{algorithm}

\section{Results and Discussion}
This section presents the simulation results of TASAC based batch transesterification process control.
The efficacy of the proposed TASAC based controller is verified by comparing it with state-of-art SAC and DDPG algorithms for three scenarios: i) Performance under nominal condition, ii) Performance in the presence of measurement noise and iii) under batch-to-batch variations. %All the algorithms are trained using the reward function described in Section \ref{S:3}.

%\subsection{Case Study: Batch Transesterification Process}

\subsection{System Description}

To study the effectiveness of TASAC based controller, the batch transesterification process is chosen as a case study. Transesterification is a process wherein triglycerides (TG) present in the lipids/fatty acids reacts with short-chain alcohol (methanol/ethanol) to produce fatty acid esters. The various sources of lipids/fatty acids include sources derived from plants such as soyabean oil, vegetable oil,waste cooking oil , palm oil,
 and animal fats, etc. \cite{khalizani2011transesterification,liu2008transesterification}. Usually, methanol and vegetable oil  are reacted in the presence of a catalyst to produce the desired product, fatty acid methyl ester (FAME).  The process of converting TG to FAME  involves a three-step reversible reaction given as follows:
\begin{align}
     TG + CH_{3}OH \underset{k_2}{\stackrel{k_1}{\rightleftharpoons}} DG + R_{1}COOCH_{3} \\
     DG + CH_{3}OH \underset{k_4}{\stackrel{k_3}{\rightleftharpoons}} MG + R_{2}COOCH_{3} \\
     MG + CH_{3}OH \underset{k_6}{\stackrel{k_5}{\rightleftharpoons}} GL + R_{3}COOCH_{3} 
 \end{align}
Diglycerides (DG) and Monoglycerides (MG) are the intermediate products and Glycerol (GL) is a byproduct. Here, the forward rate constants are $k1$, $k3$, $k4$ and the backward rate constants are $k2$, $k4$, $k6$ which are non-linearly dependent on the reactor temperature $(T_r)$ by Arhernius equation,  $k_{i}= ko_{i} \exp{(-E_{i}/RT_r)}$.

The kinetic model involving the mass balances on the reactants is adopted from \cite{noureddini1997kinetics,de2020constrained} and is given as:
\begin{align}
    \frac{d[TG]}{dt} &= -k_1[TG][A] + k_2[DG][E] \\
    \frac{d[DG]}{dt} &= k_1[TG][A] - k_2[DG][E] -k_3[DG][A] + k_4[MG][E] \\
    \frac{d[MG]}{dt} &= k_3[DG][A] - k_4[MG][E] -k_5[MG][A] + k_6[GL][E] \\
    \frac{d[E]}{dt} &= k_1[TG][A] - k_2[DG][E]+ k_3[DG][A] \\
    &
     - k_4[MG][E] +k_5[MG][A] - k_6[GL][E] \\
    \frac{d[A]}{dt} &= - \frac{d[E]}{dt} \\
    \frac{d[GL]}{dt} &= k_5[MG][A] - k_6[GL][E]\\
\end{align}
The desired FAME concentration is mainly affected by the reactor temperature $(T_r)$ and therefore is the control variable for this study. This is achieved by manipulating the jacket inlet temperature $(T_{jin})$ (in a jacketed batch reactor) and therefore is the control input. 
The energy balance equations of the batch reactor representing reactor temperature $T_r$ and  and the jacket temperature  $T_j$ are given as follows:
\begin{align}
    \frac{dT_r}{dt} &= \frac{M_R}{V\rho_{R}c_{m,R}}(-V\Delta{H}_Rr_{E}+Q_j) \\ 
    \frac {dT_j}{dt} &= \frac{F_j(T_{jin}-T_{j})}{V_j \rho_{j}}- \frac{Q_j}{V_j\rho_{j} c_w} \\
    Q_j &= UA(T_j -T_r)\\
    r_{E} &= \frac{d[E]}{dt}
\end{align}
where, $\Delta{H}$ is the heat of reaction (of the combined reaction),$r_{E}$ is the reaction rate,$M_R$ is the molar mass of reactor content, $Fj$ is the mass flow rate of jacket fluid. The model equations and the values of the parameter are taken from  \cite{chanpirak2017improvement,kern2015advanced}. 

\subsection{TASAC training details}

TASAC and other bench-marking algorithms are implemented in Python 3.7, and the deep neural networks are trained using the  Pytorch package in Python. The process's mathematical model is simulated in MatLab and integrated into Python via Matlab Engine API for Python. 

The critic and actor networks are parameterised DNNs consisting of 4 hidden layers with 512 hidden nodes. A ReLU activation function is used between each hidden layer for both actor and critic networks. The actor is a stochastic network that receives the state as input and outputs the mean and standard deviation. The critic takes in both the state and action and returns the Q-value. In addition, the Adam optimiser and the TD-error are used to update the network's weights. The hyperparameters used for training the algorithm are presented in Table\ref{Hyperparameters}. The state is two dimensional tuple, consisting of error $e(t)=(T_r - T_{ref})$, and time $t$, $s(t)= \transpose{[e(t),t]}$ and jacket inlet temperature $(a = T_{j,in})$ is the control input.

\begin{table}{H}
\centering
\caption{Hyperparameters used in DNNs : TASAC}
\label{Hyperparameters}
\begin{tabular}{lllll}
\hline
Hyperparameters & Value 
\\
\hline
Discount factor($\gamma$)& 0.99\\
Mini batch size & 100\\
Actor Learning Rate & 3$\times$10e-4\\
Critic Learning Rate & 3$\times$10e-4\\
Entropy Learning Rate & 3$\times$10e-4\\
Target Update coefficient($\tau$) & 0.01 \\
 \\
\hline
\end{tabular}
\end{table}

\subsection{Reward function}
In this study, we have used the reward function shown in (\ref{reward}) as the objective here is to minimize the tracking error scaled by time. Thus, a higher penalty is given to the steady-state tracking error. 
\begin{equation}
    r_t= -\abs{e(t)}t
    \label{reward}
\end{equation}
 where error $e(t):x(t)-x_{ref} \in \mathbb{R}^n$.

\subsubsection{ Performance Comparison}
The proposed controller is tested for temperature control in the batch transesterification process, and the performance was compared with vanilla SAC and DDPG algorithms. 
The TASAC based controller is trained for 100 episodes, and the average integral time absolute error (ITAE) value for the last 10 episodes is calculated for all the five case studies . The ITAE is considered for performance comparison as the instantaneous reward function is the magnitude of the error scaled by time as given in equation \ref{reward} for all the algorithms. A total of 10 different seeds were trained for the TASAC controller.
The average ITAE value of the 10 random seeds is reported in Table \ref{TASAC Performance studies}.  The values shows that Study1, which is the min-min case, has the least ITAE value of 409.99 as compared to other cases. The combination of max-min and max-max gives poor performance indicating the overestimation of value function.  Since the average ITAE value for the min-min combination is smaller when compared to all other cases, further analysis is done only for this case.

\begin{table}
\centering
\caption{Comparison of controller performance for different studies for TASAC algorithm (for 10 different random seeds)  }
\label{TASAC Performance studies}
\begin{tabular}{lllll}
\hline
Study & Average (ITAE) 
\\
\hline
min-min & 409.99  \\
min-max &  422.07 \\
max-min & 1313.23 \\
max-max & 925.19  \\
min-avg & 437.25  \\
\hline
\end{tabular}
\end{table}

\begin{figure}
        \centering
        \subfloat[ TASAC based controller]
        {\includegraphics[width=0.8\linewidth]{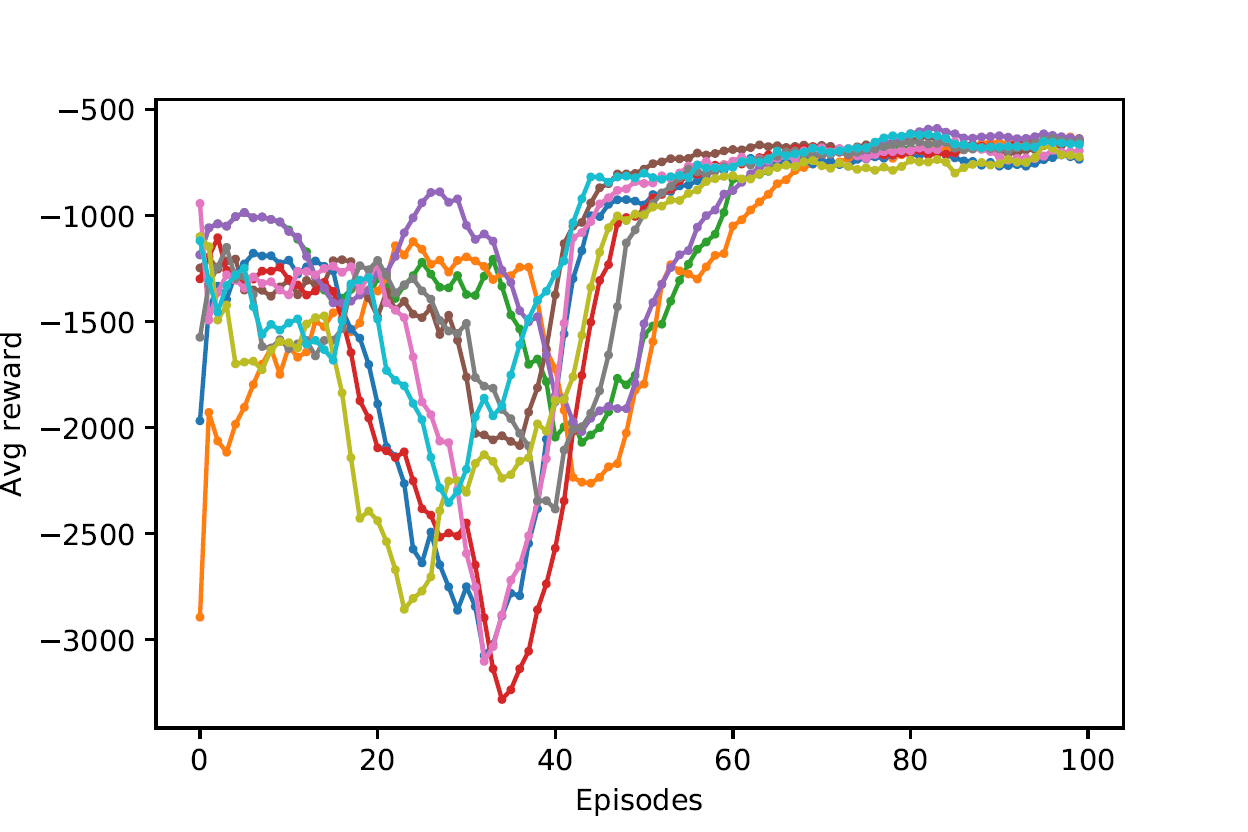}}\\
        \subfloat[SAC based controller]
        {\includegraphics[width=0.8\linewidth]{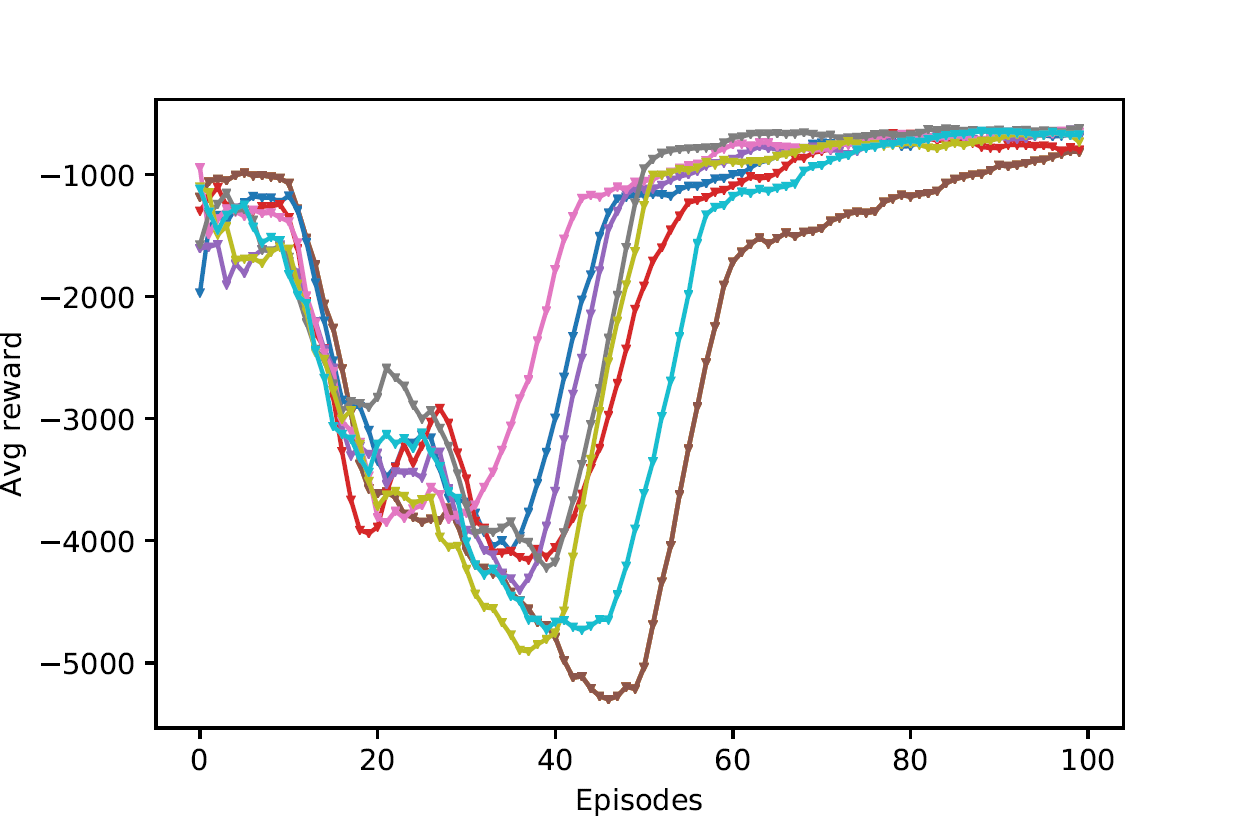}}
        \caption{Average rewards for 10 different seeds a.) TASAC based controller b.) SAC based controller.   TASAC based controller achieves better reward when compared to SAC based controller.}
        \label{fig:avg_reward_plots}
\end{figure}

Figure \ref{fig:avg_reward_plots} (a) shows the average of the last 10 episodic rewards for the min-min case for 10 different seeds of the TASAC algorithm. 
To show the effectiveness of the TASAC algorithm, the average reward plot (for 10 seeds) obtained in vanilla SAC is shown in Figure  \ref{fig:avg_reward_plots} (b).
It can be seen from Figure \ref{fig:avg_reward_plots} that the average reward obtained in TASAC having twin actors is higher than the reward obtained in SAC.

%\begin{figure}
        %\centering
        %\includegraphics[width=0.8\linewidth]{Figures/avg_rewards1A_modified_seed_itae_final.pdf}
        %\caption{Average rewards for 10 different seeds (SAC controller)}
        %\label{fig:avg_reward_SAC}
%\end{figure}

To further illustrate the proposed algorithm's effectiveness, a comparative analysis between the proposed TASAC algorithm and the vanilla SAC algorithm and DDPG was implemented. Table \ref{Controller Performance} shows the tracking error (in terms of ITAE values) between $T_r$ and $T_{ref}$ for TASAC and SAC algorithm. Here, the ITAE value reported is the average of the last 10 batches. It is clear from the values that the TASAC has a lower ITAE value of 401.12 as compared to SAC based controller.

\begin{figure}
        \centering
        \includegraphics[width=0.8\linewidth]{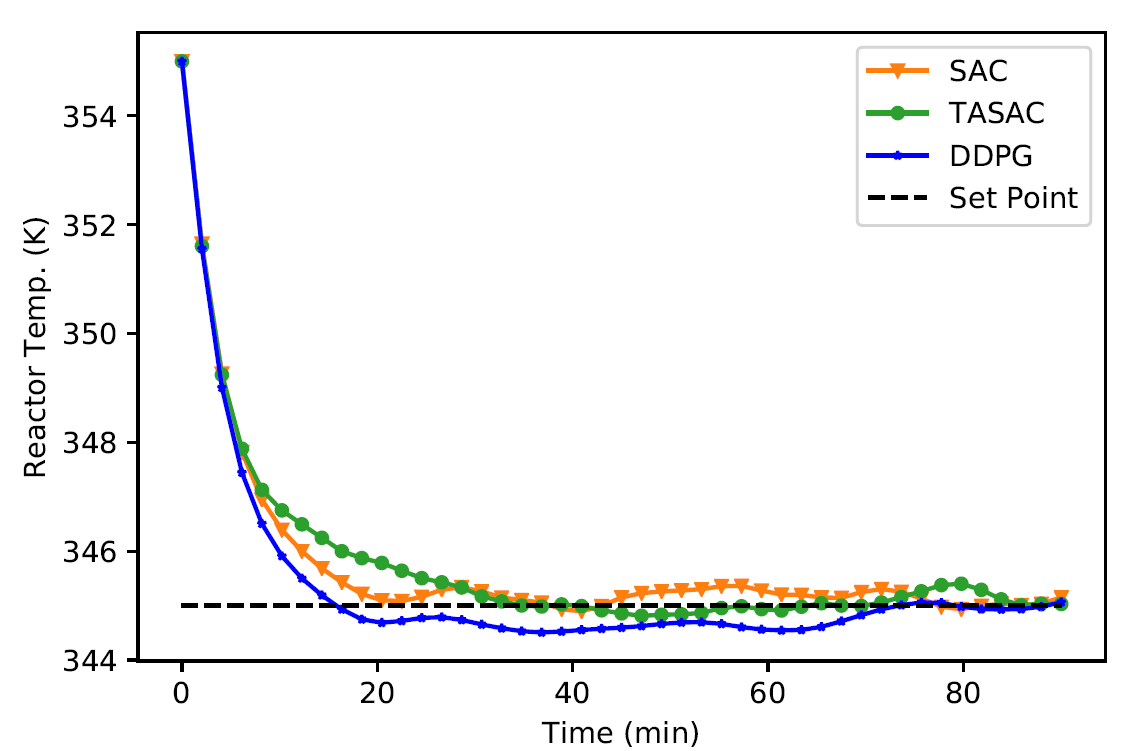}
        \caption{Comparison of tracking performance of TASAC,SAC and DDPG controller }
        \label{fig:Tr_deterministic}
\end{figure}

\begin{table}
\centering
\caption{Comparison of controller performance for TASAC $\&$ SAC algorithm (under nominal conditions) }
\label{Controller Performance}
\begin{tabular}{lc}
\hline
Algorithm & Tracking Error (ITAE) 
\\
\hline
TASAC & 401.12  \\
SAC &  414.82\\
DDPG & 541.08 \\

\hline
\end{tabular}
\end{table}
Figure \ref{fig:Tr_deterministic} shows the comparison of the tracking performance (in terms of ITAE) of the reactor temperature $(T_r)$. It is clear from Figure \ref{fig:Tr_deterministic} that TASAC is able to track the reactor temperature better as compared to SAC showcasing its superiority.

\subsection{Performance in the presence of measurement noise}
\begin{figure}
        \centering \includegraphics[width=0.8\linewidth]{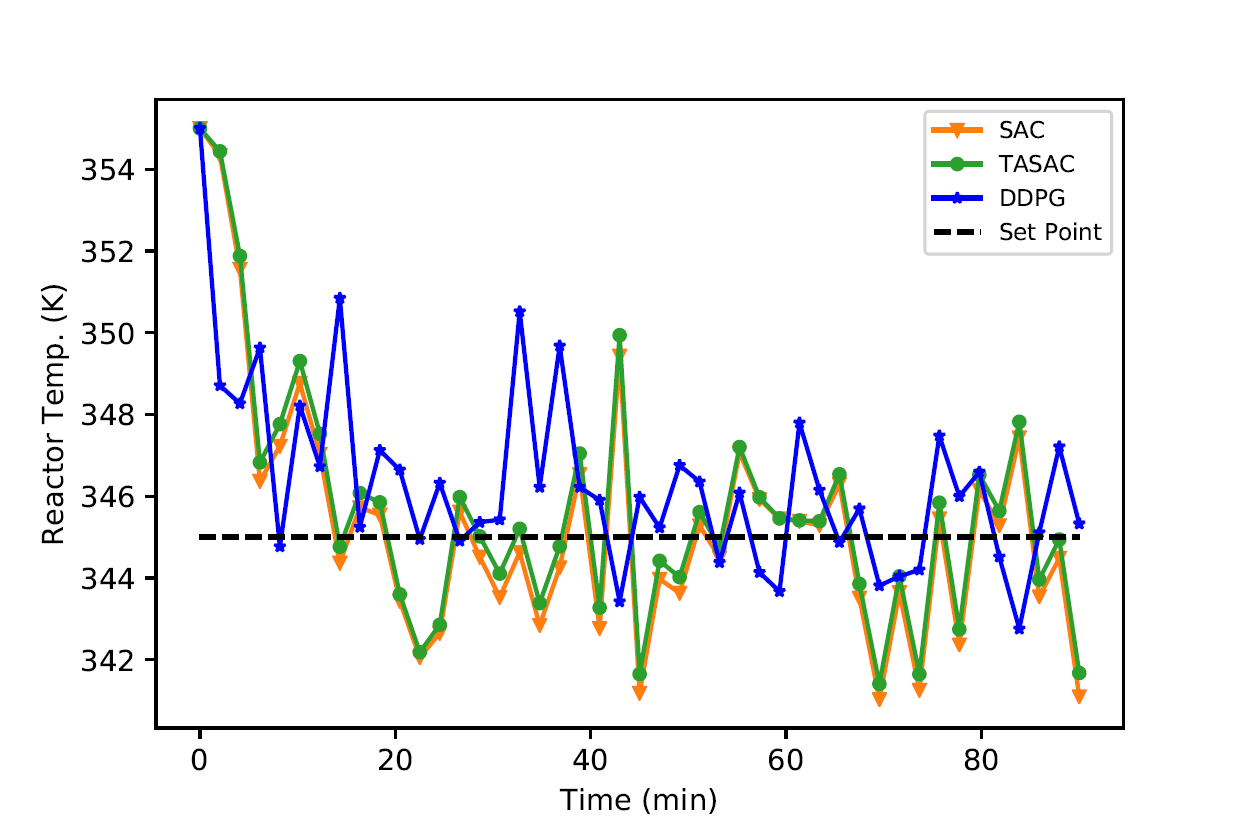}
        \caption{Comparison of tracking performance of TASAC, SAC and DDPG controller(measurement noise)  }
        \label{noise_graph}
\end{figure}

To evaluate the performance of TASAC in presence of noisy data, measurement noise is introduced in  reactor temperature by varying the state from its mean value by 0.5 percent. The average ITAE values of the last 20 batches  are reported in Table \ref{Controller Performance_noise} and compared with vanilla SAC and  DDPG algorithm, which uses deterministic policy. It can be seen from the ITAE values that TASAC has the lowest ITAE value as compared to other algorithms. The corresponding graph for the controller performance is shown in Figure \ref{noise_graph} Thus, the TASAC controller, owing to the presence of twin actors and stochastic policy, robustly control the system to the set target value even in the presence of measurement uncertainties.

\begin{table}
\centering
\caption{Comparison of controller performance for TASAC and SAC algorithm in the presence of measurement noise }
\label{Controller Performance_noise}
\begin{tabular}{lc}
\hline
Algorithm & Tracking Error(ITAE) 
\\
\hline
TASAC & 744.66  \\
SAC & 1057.20 \\
DDPG & 1434.35 \\
\hline
\end{tabular}
\end{table}

\subsection{Effect of batch-to-batch variations}
\begin{figure}
        \centering
        \includegraphics[width=0.8\linewidth]{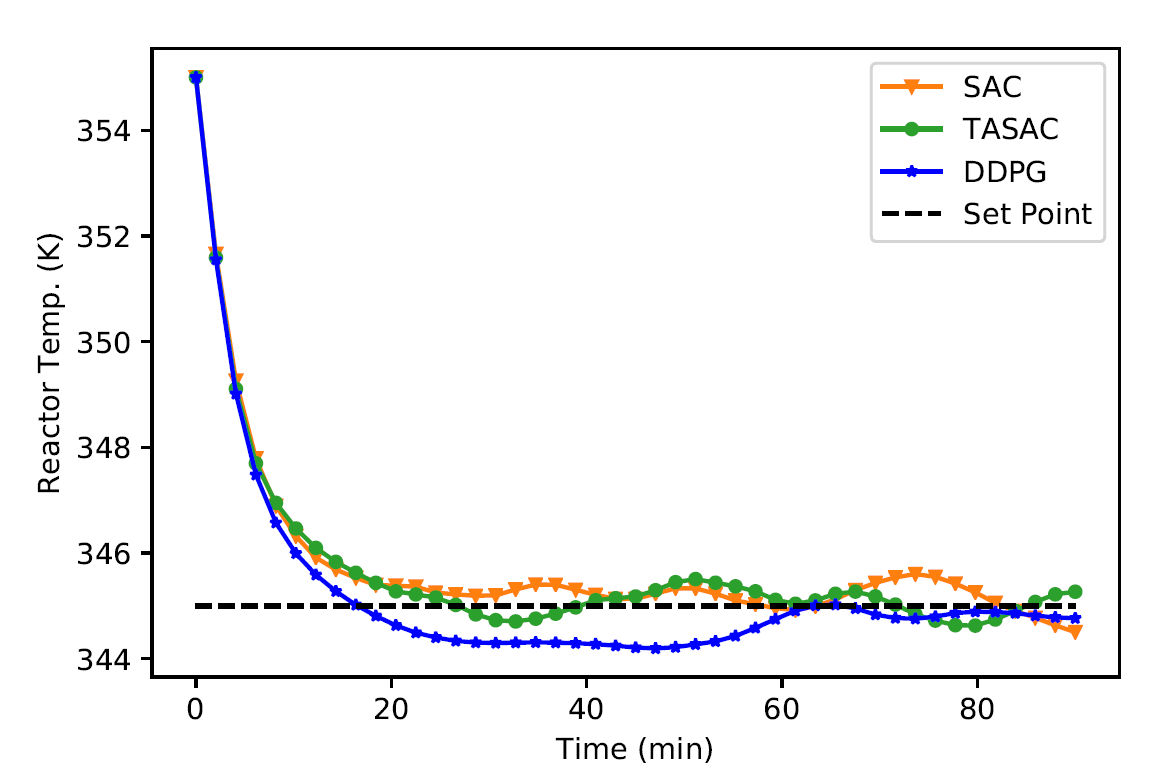}
        \caption{Comparison of tracking performance of TASAC, SAC and DDPG controller(batch-to-batch variation)  }
        \label{fig:btbv}
\end{figure}

Variations in operating conditions in different batches due to changes in raw material composition, perturbations in process parameters and external environmental conditions result in batch-to-batch variations (btbv).
In this study, btbv are introduced  by randomly changing
the rate constant, $k_c$, by changing the pre-exponential factor ($k_o$)
by ten per cent in each batch. 
The proposed controller's performance is compared to that of the DDPG and SAC algorithms, with the ITAE values reported in Table \ref{ControllerPerformance_btbv}.
The ITAE values shown in Table \ref{ControllerPerformance_btbv} indicates the average ITAE value of the last 10 batches. 
The tracking trajectory plot in Figure \ref{fig:btbv} clearly illustrates that the TASAC controller is capable of  achieving the desired control performance by reaching the set-point, $T_ref$, even in the presence
of btbv.

\begin{table}
\centering
\caption{Comparison of controller performance for TASAC and DDPG algorithm (under batch to batch variation) }
\label{ControllerPerformance_btbv} 
\begin{tabular}{lc}
\hline
Algorithm & Tracking Error(ITAE) 
\\
\hline
TASAC & 455.56  \\
SAC & 555.05 \\
DDPG & 790.07  \\
\hline
\end{tabular}
\end{table}

\begin{figure}
        \centering
        \includegraphics[width=0.8\linewidth]{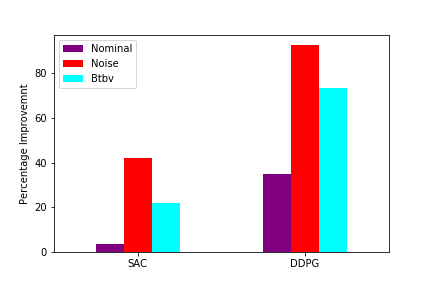}
        \caption{Percentage improvement in ITAE using TASAC comparing with SAC and DDPG}
        \label{fig:improve}
\end{figure}

Figure \ref{fig:improve} shows the percentage improvement in ITAE brought out by TASAC for all the three cases (nominal, measurement noise and batch-to-batch variation)
It is observed that the percentage improvement for TASAC  increases in the uncertain cases like batch-to-batch variations (btbv) and with measurement noise when compared to deterministic case. Figure \ref{fig:improve} clearly shows that as compared to vanilla SAC, TASAC controller shows 21.8 $\%$  improvement for btbv and 42 $\%$ in the presence of measurement noise. Further, as compared to DDPG, 73.4 $\%$  improvement is observed for btbv case and 92 $\%$  in the case of measurement noise. 
This shows that TASAC handles uncertainty and noise better when compared to SAC and DDPG.

\section{Conclusion}
A stochastic actor-critic RL based control algorithm, termed as TASAC,  is proposed in the work, with emphasis  on the twin actor networks to further enhance the   exploration ability. The proposed algorithm is validated by implementing it  for the control of reactor temperature, in batch transesterification process. In addition, a decision making criteria to select the best action from the two actor networks outputs is proposed, considering the over estimation of the value function. Compared to SAC algorithm, the proposed controller showed improved performance by reaching the target set-point value both in i) the nominal conditions ii) presence of  measurement noise and iii) when batch-to-batch variations are introduced in the process. Compared to the DDPG which has a deterministic policy, TASAC based controller incorporated twin actors and stochastic policy, showed enhanced ability in tracking performance in the presence of batch-to-batch variations and measurement uncertainties. The results indicates that TASAC based controller can be a potential framework to achieve the goal of AI based control in process industries.
\section*{Acknowledgements}
The authors gratefully acknowledge the funding received from SERB India with the file number CRG/2018/001555 and TiHAN, IIT Hyderabad with TiHAN-IITH/03/2021-22/34(4).
\bibliographystyle{elsarticle-num-names}
\bibliography{sample.bib}






% \end{thebibliography}

\end{document}